\pdfoutput=1
\documentclass[11pt]{article}

\usepackage[preprint]{acl}

\usepackage{times}
\usepackage{latexsym}
\usepackage{bbding}
\usepackage{colortbl}
\usepackage{graphicx}
\usepackage{multirow}
\usepackage{booktabs}
\usepackage{hhline}
\usepackage{amsmath}
\usepackage{amssymb}
\usepackage{pifont}
\usepackage{algorithm}
\usepackage{algorithmic}
\usepackage{arydshln}
\usepackage{subfigure}
\usepackage{url}
\usepackage{enumitem}
\usepackage[utf8]{inputenc}

\usepackage{cleveref}
\crefname{section}{§}{§§}
\Crefname{section}{§}{§§}

\usepackage[T1]{fontenc}
\usepackage[utf8]{inputenc}

\usepackage{microtype}

\usepackage{inconsolata}

\title{Sparsity-Accelerated Training for Large Language Models}

\renewcommand{\thefootnote}{\fnsymbol{footnote}}

\author{Da Ma\footnotemark[4], Lu Chen\footnotemark[4]\footnotemark[2]\footnotemark[1], Pengyu Wang\footnotemark[4]\footnotemark[2], Hongshen Xu\footnotemark[4], Hanqi Li\footnotemark[4] \\ \textbf{Liangtai Sun}\footnotemark[4], \textbf{Su Zhu}\footnotemark[3], \textbf{Shuai Fan}\footnotemark[3], \textbf{Kai Yu}\footnotemark[4]\footnotemark[2]\footnotemark[1]\\
  \footnotemark[4]X-LANCE Lab, Department of Computer Science and Engineering \\
  MoE Key Lab of Artificial Intelligence, SJTU AI Institute \\
  Shanghai Jiao Tong University, Shanghai, China \\
  \footnotemark[2]Suzhou Laboratory, Suzhou, China\\
  \footnotemark[3]AISpeech Co., Ltd., Suzhou, China\\
  \texttt{\{mada123, chenlusz, kai.yu\}@sjtu.edu.cn} \\ 
  }

\begin{document}

\maketitle
\footnotetext[1]{The corresponding authors are Lu Chen and Kai Yu.}

\renewcommand{\thefootnote}{\arabic{footnote}}

\begin{abstract}
Large language models (LLMs) have demonstrated proficiency across various natural language processing (NLP) tasks but often require additional training, such as continual pre-training and supervised fine-tuning. However, the costs associated with this, primarily due to their large parameter count, remain high. This paper proposes leveraging \emph{sparsity} in pre-trained LLMs to expedite this training process. By observing sparsity in activated neurons during forward iterations, we identify the potential for computational speed-ups by excluding inactive neurons. We address associated challenges by extending existing neuron importance evaluation metrics and introducing a ladder omission rate scheduler. Our experiments on Llama-2 demonstrate that Sparsity-Accelerated Training (SAT) achieves comparable or superior performance to standard training while significantly accelerating the process. Specifically, SAT achieves a $45\%$ throughput improvement in continual pre-training and saves $38\%$ training time in supervised fine-tuning in practice. It offers a simple, hardware-agnostic, and easily deployable framework for additional LLM training. Our code is available at \url{https://github.com/OpenDFM/SAT}.
\end{abstract}
\label{sec:introduction}
\section{Introduction}

% Introduce additional training
Large language models~(LLMs), such as GPT-4~\cite{achiam2023gpt}, Mistral~\cite{jiang2023mistral}, and Llama-2~\cite{touvron2023llama}, have demonstrated 
\begin{figure}[t]
    \centering
    \includegraphics[width=\linewidth]{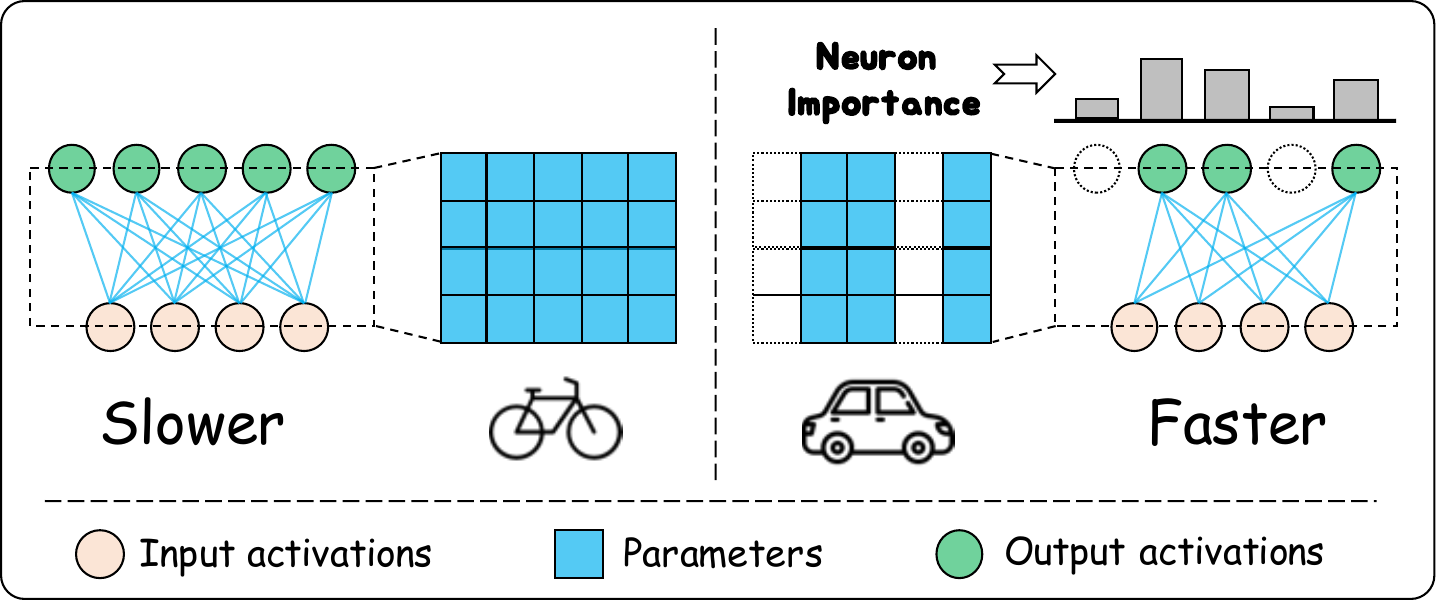}
    \caption{Insights into neuron sparsity in large language models to accelerate training. \textbf{Left}: standard linear layer. \textbf{Right}: the linear layer after pruning relatively less important neurons. By omitting computations related to some less important neurons, training on the right side proceeds faster compared to the left side, while maintaining similar performance.}
    \label{fig:introduction}
\end{figure}
remarkable capabilities across numerous NLP tasks~\cite{mao2023gpteval,dillmann2024evaluation}.
In general, all these models are initially pre-trained on massive data by unsupervised learning. Furthermore, such models regularly necessitate additional training for two primary scenarios: 1) \emph{continual pre-training} on new data as the pool of available pre-training data continuously expands~\cite{gupta2023continual,cui2023efficient,ke2023continual}. 2) \emph{supervised fine-tuing}~(SFT) on labeled data to enhance the capacity of LLMs to follow human instructions~\cite{ouyang2022training,dong2023abilities,ansell2024scaling}.

% However, the pool of available pre-training data is continuously expanding~\cite{gupta2023continual}. Retraining LLMs from scratch on the combined existing and new data is impractical due to the high costs involved.

% State the time\&computation consuming problem
In contrast to initial pre-training, the additional training demands relatively less time and computational resources. Regrettably, it remains somewhat challenging to afford, primarily attributable to the substantial parameter count~\cite{zhao2024chemdfm,wu2023pmc}. Fortunately, numerous practitioners~\cite{zhao2024apt} are dedicating efforts to saving such additional training costs. However, they mainly focus on the Parameter-Efficient Fine-Tuning~(PEFT) methods~\cite{ansell2024scaling,lialin2023scaling}, which is not commonly employed in the aforementioned continual pre-training~\cite{cui2023efficient} scenario. Hence, it is imperative to explore novel acceleration methods suitable for both additional training scenes simultaneously.

% Introduce sparsity in LLMs and describe our insights.
In this paper, our objective is to leverage the structural \emph{sparsity}~\cite{li2023the} inherent in pre-trained LLMs to expedite their additional training scenarios. In detail, researchers~\cite{zhang-etal-2022-moefication,liu2023deja} have observed that the activated~(important) neurons\footnote{In this paper, a neuron is associated with a specific row/column in a weight matrix.} exhibit sparsity in each forward iteration. Intuitively, we can speed up by omitting the calculation of inactive~(unimportant) neurons. As depicted in Figure \ref{fig:introduction}, suppose a neuron corresponds to a particular column of the weight matrix in a linear layer. In each training iteration, a \emph{compact} and \emph{efficient} neuron importance assessment module~(right part) would evaluate the activation status of neurons in advance. By discarding those inactive neurons~(columns in the weight matrix), training speed is enhanced.

% Throw potential problems behind such a method and elaborate our work to tackle the problems
Aside from the enhancement in computational speed, disregarding the computation of inactive neurons also poses the following three additional challenges: 1) determining the method of neurons to discard, 2) assessing whether omitting neurons harms the model performance, and 3) evaluating the effectiveness of the approach in real additional training scenarios of LLMs. For the first challenge, we borrowed and extended several famous neuron importance evaluation metrics~\cite{liu2023deja,han2015learning,sun2023simple} within the context of transformer pruning, and unimportant neurons are dropped. Extensive experiments~(\cref{subsec:exp-ns}) on TinyLlama-1.1B~\cite{zhang2024tinyllama} help us determine~(\cref{subsec:SAT-for-trans}) 1) using maxip~(a neuron importance metric) to assess neuron importance, 2) adopting sampling~(versus top-k) neuron selection strategy, and 3) introducing a ladder omission rate scheduler. 

For the latter two, we separately conduct experiments over Llama-2~($7$B and $13$B) in both standard and sparsity-accelerated manners for both continual pre-training and supervised fine-tuning scenarios~(\cref{subsec:exp-performance}). Evaluation results on several benchmarks show that SAT achieves comparable sometimes even better performance compared to normal training. Moreover, it can obtain about $45\%$ throughput improvement for continual pre-training and $38\%$ speedup in elapsed time for supervised fine-tuning~(\cref{subsec:exp-speedup}). In summary, this work contributes in the following three aspects:
\begin{itemize}
    \item[1.] We propose the \textbf{S}parsity-\textbf{A}ccelerated \textbf{T}raining~(SAT), a novel, hardware-agnostic, and easily deployable framework for additional training of LLMs.
    \item[2.] We investigate several neuron importance computation and selection methods and summarize an optimal configuration for SAT. In addition, a novel ladder omission rate scheduler is designed to alleviate the overfitting problem of SAT.
    \item[3.] Extensive experiments in both continual pre-training and supervised fine-tuning demonstrate that SAT achieves comparable performance to standard training of LLMs with speedup.
\end{itemize}

\label{sec:method}
\section{Methodology}
In this section, we first give an overview of SAT. Next, more details are provided within the context of transformers, which is the predominant architecture of recent LLMs. Finally, we discuss our implementation and conduct a theoretical efficiency analysis for the SAT.

\subsection{Overview of SAT}
Suppose a neural network model is expressed by $f(\cdot;\mathcal{N}_{\theta})$, where $\mathcal{N}_{\theta}$ denotes the set of neurons. Recall that our objective is to disregard non-essential neurons to accelerate. Consequently, Sparse-Accelerated Training (SAT) can be delineated into the following two overarching steps for the $t$-th training iteration:
\begin{itemize}
    \item \emph{Neuron Importance Computation}: compute the importance scores for each neuron,
    \begin{align}
        \mathbf{s}_\theta^t &= \textit{ComputeImportance}\left(\cdot, \mathcal{N}_{\theta}^t\right),
    \end{align}
    where $\mathbf{s}_\theta^t\in\mathbb{R}^{|\mathcal{N}_{\theta}^t|}$.
    \item \emph{Neuron Selction and Optimization}: select important neurons upon $\mathbf{s}_\theta^t$ and update them,
    \begin{equation}
        \begin{aligned}
            \tilde{\mathcal{N}_{\theta}^t} =& \textit{SelectNeuron}\left(\mathbf{s}_\theta^t, r, \mathcal{N}_{\theta}^t\right),\\
            \mathcal{N}_\theta^{t+1} =& \left\{n_\theta\middle| n_\theta\in\left(\mathcal{N}_\theta^t-\tilde{\mathcal{N}_\theta^t}\right) \right. \\
            & \left. \vee n_\theta \in \textit{Optimize}\left(\tilde{\mathcal{N}_\theta^t}\right)\right\},
        \end{aligned}
    \end{equation}
    where $r\in[0, 1)$ denotes the neuron omission rate and $\tilde{\mathcal{N}_{\theta}^t}\subseteq\mathcal{N}_{\theta}^t$. Actually, we optimize the subnetwork expressed by $f(\cdot; \tilde{\mathcal{N}_{\theta}^t})$ at training step $t$.
\end{itemize}

\subsection{SAT for transformers}
\label{subsec:SAT-for-trans}
As widely recognized, contemporary LLMs are predominantly based on the Transformer~\cite{vaswani2017attention} architecture. Therefore, our study primarily explores the SAT for transformers.

\paragraph{Review of Transformers} The transformer is composed of numerous stacked layers, with each layer consisting of two main modules: Multi-Head Attention~(MHA) and an immediately subsequent Multi-Layer Perceptron~(MLP). Formally, let $\mathbf{X}=[\mathbf{x}_1,\mathbf{x}_2,\ldots,\mathbf{x}_n]\in\mathbb{R}^{n\times d}$ denote the input activations of the $\ell$-th layer\footnote{To maintain manuscript tidiness, we omit the subscript of $\ell$ in this paper.}. The MHA can be formulated as
\begin{equation}
    \begin{gathered}
        \label{formula:MHA}
        \textit{MHA}\left(\mathbf{X}\right) = \textit{Concat}\left(\textit{head}_1,\ldots,\textit{head}_h\right)\mathbf{W}_\textit{O},\\
        \textit{head}_i = \mathbf{A}_i\cdot\left(\mathbf{X}\mathbf{W}_\textit{V}^i\right), \\
        \mathbf{A}_i = \textit{Softmax}\left(\frac{\left(\mathbf{X}\mathbf{W}_\textit{Q}^{i}\right)\cdot\left(\mathbf{X}\mathbf{W}_\textit{K}^i\right)^T}{\sqrt{d_k}}\right),
    \end{gathered}
\end{equation}
where $h\in\mathbb{N}^+$ represents the number of heads and $d\in\mathbb{N}^+$ is the dimension of hidden states. $d_k=d/h$,  $\mathbf{W}_\textit{Q}^i,\mathbf{W}_\textit{K}^i$ and $\mathbf{W}_\textit{V}^i\in\mathbb{R}^{d\times d_k}$~($1\leq i \leq h$), and $\mathbf{W}_\textit{O}\in\mathbb{R}^{d\times d}$ are trainable parameters.

Subsequently, the MLP takes the output of MHA with a residual connection as input. Mathematically,
\begin{equation}
    \begin{gathered}
    \mathbf{X} = \mathbf{X} + \textit{MHA}\left(\mathbf{X}\right), \\
    \label{formula:MLP}
    \textit{MLP}\left(\mathbf{X}\right) = \sigma\left(\mathbf{X}\mathbf{W}_\textit{up}\right)\cdot\mathbf{W}_\textit{down},
    \end{gathered}
\end{equation}
where $\sigma\left(\cdot\right)$ is the activation function, $\mathbf{W}_\textit{up}\in\mathbb{R}^{d\times 4d}$ and $\mathbf{W}_\textit{down}\in\mathbb{R}^{4d\times d}$ are trainable parameters. The output of MLP with a residual connection is fed into the next layer\footnote{We omit the writing of layer norm layer and attention mask here.}. Next, we shall proceed to a detailed discussion of the two steps of SAT within the Transformer architecture.

% rewrite with theoretical explanation
\paragraph{Neuron Importance Computation}
In this work, we leverage the structural sparsity of neurons across all linear layers, i.e., each row or column of the parameter matrix is conceptualized as a neuron. As our approach solely depends on the significance of columnar neurons, we exclusively focus on devising the methodology for assessing their importance. To clarify these methods, we first introduce some notations. Assume a column neuron $\mathbf{v}\in\mathbb{R}^{d\times 1}$ and it takes $\mathbf{Z}\in\mathbb{R}^{m\times d}$ as input. $\mathbf{Z}_i$ and $\mathbf{Z}^j$ are the $i$-th row and $j$-th column of $\mathbf{Z}$, respectively. $s_\mathbf{v}$ denotes the neuron importance score of $\mathbf{v}$. The activation value of $\mathbf{v}$ is $\mathbf{y}=\mathbf{Z}\cdot \mathbf{v}$. Next, we first mathematically define four evaluation metrics~\cite{liu2023deja,han2015learning,sun2023simple} for neuron importance, which are
\begin{equation}
\begin{aligned}
    \textit{uniform}& : s_\mathbf{v} = 1, \\
    \textit{magnitude}& : s_\mathbf{v} = \left\|\mathbf{v}\right\|_2, \\
    \textit{wanda}& : s_\mathbf{v}=\Big[\left\|\mathbf{Z}^1\right\|_2,\ldots,\left\|\mathbf{Z}^d\right\|_2\Big]\cdot \mathbf{v}, \\
    \textit{maxip}& : s_\mathbf{v}=\left(\sum\limits_{i=1}^m\mathbf{Z}_i\big/m\right)\cdot\mathbf{v}.
\end{aligned}
\end{equation}
The \emph{uniform} method is straightforward, as it entails randomly preserving neurons, akin to the well-known vanilla dropout technique~\cite{DBLP:journals/corr/abs-1207-0580}. The other three methods are predicated on the assumption that \emph{neurons with larger activation values are relatively more important}~\cite{liu2023deja}. To elucidate this assumption, we can consider an example. In the multi-head self-attention mechanism of the Transformer, the representation of each token is weighted by the representations of all input tokens. At this juncture, the weights (attention scores) between tokens become pivotal. These attention scores stem from the dot product of the representations (activation values) of each token. The higher the activation value, the greater the probability of a larger dot product, yielding a higher attention score, thus indicating greater importance to the token representation.

% explain the other three methods
Based on this assumption, the remaining three methods posit that neurons more likely to generate higher activation values~($\mathbf{y}$) are more important. In detail, the \emph{magnitude}~\cite{han2015learning} method solely focuses on the magnitude of the neuron parameter, i.e., the L2 norm of $\mathbf{v}$. The remaining two methods consider the influence of both input and parameters on the activation values. Concretely, \emph{wanda}~\cite{sun2023simple} considers the magnitude of the input and parameter product, while \emph{maxip}~\cite{liu2023deja,song2023powerinfer} focuses on the input mean and parameter product.

Subsequently, we will delve into separate discussions regarding neuron selection for MHA and MLP according to the neuron importance score.

\begin{figure*}[t]
    \centering
    \includegraphics[width=\textwidth]{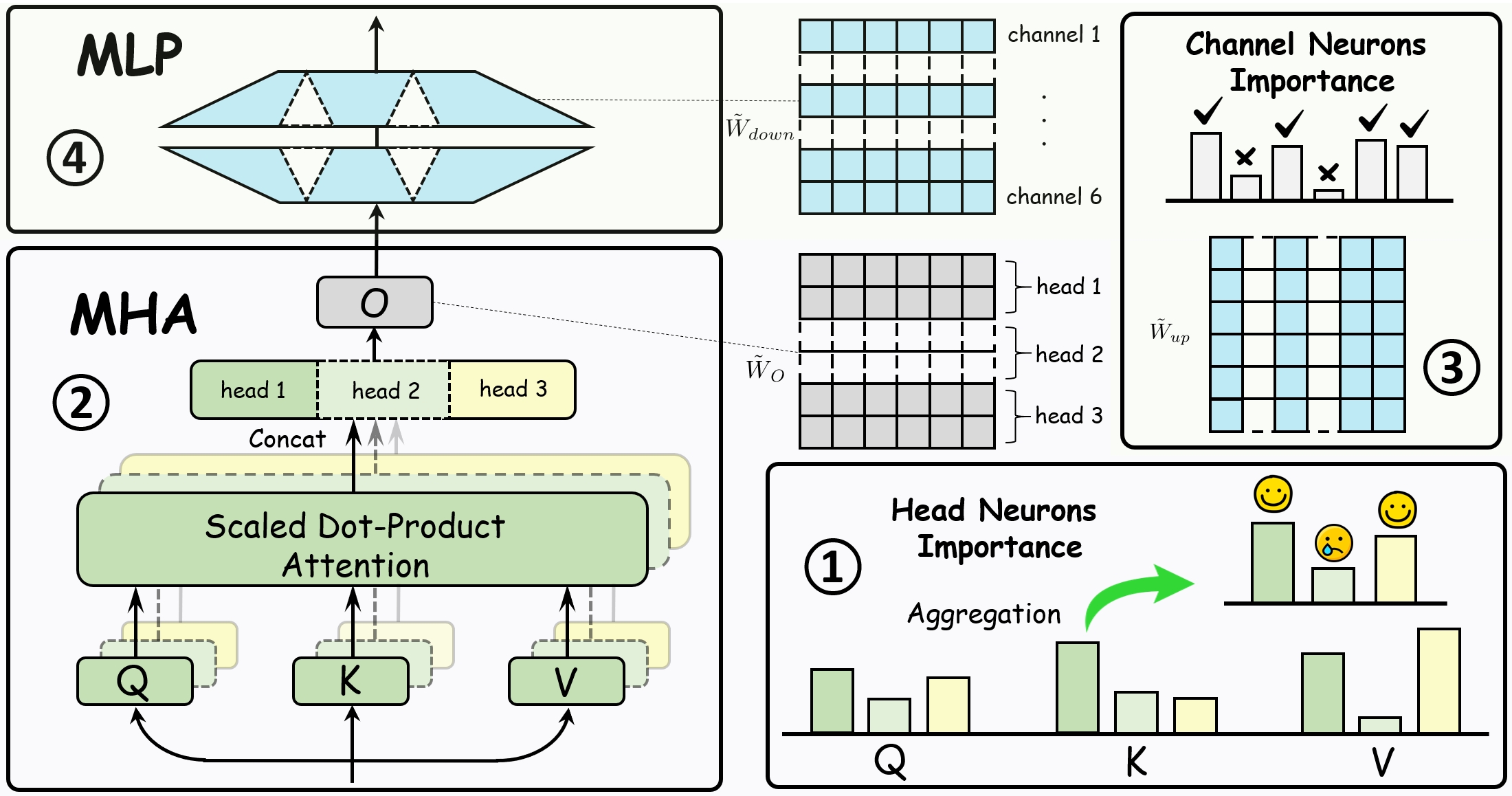}
    \caption{Sparsity-Accelerated Training~(SAT) for transformers. \ding{172}: Process of selecting more important heads for MHA. \ding{173}: SAT for MHA. \ding{174}: Process of selecting more important channels for MLP. \ding{175}: SAT for MLP. Dashed parts denote omitted neurons.} 
    \label{fig:model}
\end{figure*}

\paragraph{Neuron Selection for MHA} For MHA, we operate at the granularity of individual heads, where we either omit or retain all neurons within a head. For the $i$-th head, each of its three linear parameter matrices $\mathbf{W}_\textit{Q}^i$, $\mathbf{W}_\textit{K}^i$, $\mathbf{W}_\textit{V}^i\in\mathbb{R}^{d\times d_k}$ includes $d_k$ neurons, and the neuron importance score of the $j$-th neuron in each matrix is denoted as $s_\textit{Q}^{i,j}$, $s_\textit{K}^{i,j}$, $s_\textit{V}^{i,j}$ respectively. We first calculate the importance score of the $i$-th head in each parameter matrix via
\begin{equation}
\begin{gathered}
s_{\textit{Q}}^{\textit{head}_i}=\sum\limits_{j=1}^{d_k}\frac{s_\textit{Q}^{i,j}}{d_k},
s_{\textit{K}}^{\textit{head}_i}=\sum\limits_{j=1}^{d_k}\frac{s_\textit{K}^{i,j}}{d_k},
s_{\textit{V}}^{\textit{head}_i}=\sum\limits_{j=1}^{d_k}\frac{s_\textit{V}^{i,j}}{d_k}.
\end{gathered}
\end{equation}
Then, the importance score of $i$-th head is aggregated, as depicted in Figure \ref{fig:model}~(\ding{172}) by 
\begin{align}
    s^{\textit{head}_i}=\left(s_{\textit{Q}}^{\textit{head}_i}+s_{\textit{K}}^{\textit{head}_i}+s_{\textit{V}}^{\textit{head}_i}\right)\Big/3.
\end{align}
Subsequently, $\tilde{h}=\left\lfloor h\times \left(1 - r\right) \right\rfloor$ heads will be selected to optimize based on the head importance score in the following two ways: 
\begin{itemize}
    \item \emph{top-k}: retain the top $\tilde{h}$ heads with the highest importance scores.
    \item \emph{sampling}: sample $\tilde{h}$ heads according to the distribution:
    \begin{equation}
    \begin{gathered}
        \textit{head}_i \sim p\left({\textit{head}_i}\right),\\
        p\left({\textit{head}_i}\right) = \frac{\textit{exp}\left(s^{\textit{head}_i}/\tau\right)}{\sum_{j} \textit{exp}\left(s^{\textit{head}_j}/\tau\right)},
        \end{gathered}
    \end{equation}
    where $1\leq i \leq h$ and $\tau\in\mathbb{R}^{+}$ is the temperature for sampling~\cite{dupont2022extracting}.
\end{itemize}
As is shown in Figure \ref{fig:model}~(\ding{173}), only selected $\tilde{h}$ heads participate in the computation. Moreover, only the rows corresponding to the selected heads in $\mathbf{W}_{\textit{O}}$~(referred to as $\tilde{\mathbf{W}}_{\textit{O}}$) will be involved in the computation. Assuming heads $(\pi_1, \pi_2, \ldots, \pi_{\tilde{h}})$ are selected, the formula (\ref{formula:MHA}) becomes
\begin{align}
    \textit{MHA}\left(\mathbf{X}\right) &= \textit{Concat}\left(\textit{head}_{\pi_1},\ldots,\textit{head}_{\pi_{\tilde{h}}}\right)\tilde{\mathbf{W}}_\textit{O}.
\end{align}

\paragraph{Neuron Selection for MLP} For MLP, we operate at the granularity of individual channels, where each channel~(column) in $\mathbf{W}_{\textit{up}}\in\mathbb{R}^{d\times 4d}$ corresponds to a neuron and we select $\tilde{c}=\left\lfloor 4d \cdot \left(1 - r\right) \right\rfloor$ neurons~(denoted as $\tilde{\mathbf{W}}_\textit{up}\in\mathbb{R}^{d\times\tilde{c}}$) to optimize similarly to MHA~(see Figure \ref{fig:model}~(\ding{174})). Likewise, rows corresponding to the selected channels in $\mathbf{W}_{\textit{down}}\in\mathbb{R}^{4d\times d}$~(denoted as $\tilde{\mathbf{W}}_\textit{down}\in\mathbb{R}^{\tilde{c}\times d}$) will participate in the computation as illustrated in Figure \ref{fig:model}~(\ding{175}). Then, the formula (\ref{formula:MLP}) turns into
\begin{align}
    \textit{MLP}\left(\mathbf{X}\right) &= \sigma\left(\mathbf{X}\tilde{\mathbf{W}}_\textit{up}\right)\cdot\tilde{\mathbf{W}}_\textit{down}.
\end{align}

% FIXME: add more explanations in rebuttal
\paragraph{Ladder Omission Rate Scheduler~(LORS)} Empirically, pruning some neurons may pose a risk of overfitting to the model. To mitigate this potential issue, we plan to divide the training process into two stages: 1) Train sparsely with a constant omission rate and 2) Gradually decrease the omission rate, making the model denser until it fully recovers as a dense model. Through the second stage, the overfitting problem of certain parameters in the model is alleviated to some extent.

For the second stage, we experimented with linear and cosine omission rate schedulers. In terms of model performance, it is common to employ these schedulers during training. However, due to the frequent fluctuations in omission rate, there is considerable variability in training speed, occasionally resulting in slower progress compared to training without neuron omission. Therefore, we considered reducing the frequency of changes in the omission rate by maintaining a constant rate for a short period before decreasing to the next constant rate segment. Additionally, similar to curriculum learning~\cite{bengio2009curriculum}, we gradually lengthen the period to allow the model to adapt to the changes in the omission rate.

Based on these insights, we propose a Ladder Omission Rate Scheduler. Figure \ref{fig:CPT-1.1B}~(c) visualizes an example of LORS. Mathematically,
considering a maximum neuron omission rate $r$ and total training steps $T$, supposing the neuron omission rate initiates its decrease at step $\eta$ with a ladder number denoted as $L$, the neuron omission rate $r_t$ at step $t$ can be formulated as follows:
\begin{equation}
        \begin{gathered}
	r_t=\left\{
	\begin{aligned}
		r & , & t \leq \eta \\
		\max \left\{0, r -              \frac{r}{L}  \cdot \ell_t \right\} & , & t > \eta
	\end{aligned}
	\right.
     ,
     \text{ where}\\
     \frac{2^{\ell_t-1}-1}{2^L-1}\leq \frac{t-\eta}{T-\eta} < \frac{2^{\ell_t}-1}{2^L-1} \text{ and } \ell_t\in\mathbb{N}^+.
     \end{gathered}
 \end{equation}

% training FLOPs calculation in rebuttal
\subsection{Implementation and Theoretical Efficiency Analysis}
 \paragraph{Implementation} Taking into account that we are accelerating the training of LLMs at the algorithmic level, it is imperative for SAT to be compatible with lower-level acceleration techniques, such as operator-level acceleration~(e.g., flash attention~\cite{dao2022flashattention, dao2023flashattention}) and training frameworks like DeepSpeed~\cite{rasley2020deepspeed}. To accomplish this, we employ an exceedingly simple and convenient implementation approach: for a linear layer, during each training iteration, the optimizer remains consistent with standard training, preserving a complete parameter matrix. Nevertheless, during computation, we exclusively engage the sparse portions~(several columns or rows) of the parameter matrix for calculation.

\paragraph{Efficiency Analysis}
 In theory, we compare SAT and standard training in terms of both memory consumption and FLOPs~(floating point of operations).
\begin{itemize}[leftmargin=2em]
    \item Memory: Given that the primary memory-intensive components on the GPU, namely the model parameters and optimizer states, remain consistent with those of standard training, the actual memory demand of our SAT during training theoretically should align closely with that of standard training.
    \item FLOPs: During both the forward and backward processes, we omit the computation of $r$ proportion of neurons. Consequently, by disregarding the calculation overhead of the second stage of LORS~(mentioned in subsection \ref{subsec:SAT-for-trans}), SAT can save approximately $\frac{\eta r}{T}$ FLOPs.
\end{itemize}

\begin{table*}
\centering
\resizebox{\textwidth}{!}{
\begin{tabular}{c|clc|cccccc:c} 
\hline

\hline

% ------------------------------------------------------------------
\multirow{2}{*}{\textbf{Order}} & \multirow{2}{*}{\textbf{CPT}} & \multirow{2}{*}{\textbf{NIM}} & \multirow{2}{*}{\textbf{NSM}}  & \multicolumn{6}{c:}{\textbf{Skywork PPL}~$\mathbf{\downarrow}$} & \multirow{2}{*}{\textbf{Avg.}~$\mathbf{\downarrow}$}     \\ 
\cdashline{5-10}
 & & & & \multicolumn{1}{c}{finance} & \multicolumn{1}{c}{game} & \multicolumn{1}{c}{general} & \multicolumn{1}{c}{government} & \multicolumn{1}{c}{movie} & \multicolumn{1}{c:}{technology} &  \\ 
\hline
% ------------------------------------------------------------------
\rowcolor{blue!5} \multicolumn{11}{c}{\emph{w/o sparsity}}\\
$1$ & \XSolidBrush & \textendash & \textendash  & $8.40$ & $16.95$ & $11.31$ & $13.00$ & $24.26$ & $11.24$ & $14.19$ \\
$2$ & \Checkmark & \textendash  & \textendash & $\mathbf{4.36}$ &	$\mathbf{11.39}$ &	$\mathbf{5.74}$ &	$\mathbf{5.08}$&	$\mathbf{15.54}$&	$\mathbf{7.50}$ &	$\mathbf{8.27}$\\

\rowcolor{blue!5} \multicolumn{11}{c}{\emph{w/ sparsity}}\\
$3$ & \Checkmark & uniform & \textendash  & $6.91$ & $18.25$ & $8.92$ & $8.61$ & $25.58$ & $11.41$ & $13.28$ \\
\cdashline{1-11}
$4$ &\Checkmark & magnitude & top-k & $30.92$ & $52.59$ & $33.26$ & $47.71$ & $77.75$ & $34.61$ & $46.14$\\
$5$ &\Checkmark & wanda & top-k  & $17.16$ & $32.04$ & $16.55$ & $24.62$ & $49.39$ & $21.51$ & $26.88$\\
$6$ &\Checkmark & maxip & top-k  & $20.60$ & $56.15$ & $23.55$ & $39.25$ & $78.19$ & $31.98$ & $41.62$\\
\cdashline{1-11}
$7$ &\Checkmark & magnitude & sampling  & $7.68$ &$21.19$ & $10.06$ & $9.42$ & $29.20$ & $13.27$ & $15.14$ \\
$8$ &\Checkmark & wanda & sampling  & $7.23$ & $18.07$ & $9.06$ & $8.97$ & $25.86$ & $11.45$& $13.44$\\
$9$ &\Checkmark & maxip & sampling  &$\mathbf{6.56}$ & $\mathbf{17.58}$ & $\mathbf{8.59}$ & $\mathbf{8.10}$ & $\mathbf{24.89}$ & $\mathbf{11.03}$ & $\mathbf{12.79}$\\

\rowcolor{blue!5} \multicolumn{11}{c}{\emph{w/ sparsity+LORS}}\\
$10$ &\Checkmark & uniform & \textendash  & $5.69$ & $13.51$ & $7.20$ & $6.72$ & $18.79$ & $9.31$ & $10.20$ \\     
$11$ &\Checkmark & maxip & sampling  & $\mathbf{4.87}$ & $\mathbf{12.54}$ & $\mathbf{6.35}$ & $\mathbf{5.82}$ & $\mathbf{17.25}$ & $\mathbf{8.19}$ & $\mathbf{9.17}$\\
   
% ------------------------------------------------------------------
\hline

\hline
\end{tabular}
}
\caption{Results of continual pre-training of TinyLlama-1.1B with different neuron importance metrics and neuron selection methods. CPT: continual pre-training; \Checkmark and \XSolidBrush denote continually pre-training the models and not, respectively. NIM: neuron importance metric. NSM: neuron selection method. LORS: ladder omission rate scheduler. The omission rate is set to $50\%$.}
\label{tab:ns}
\end{table*}

\label{sec:experiment}
\section{Experiments}
We conduct comprehensive experiments to address the following three questions:
\begin{itemize}
    \item[Q1:] What is the optimal configuration, including assessment of neurons importance~(across various metrics) and neurons selection ~(top-k vs. sampling), with LORS for SAT?
    \item[Q2:] Does SAT harm the performance of LLMs in both continual pre-training and supervised fine-tuning scenarios?
    \item[Q3:] What is the efficiency achieved by SAT over popular LLMs in practice?
\end{itemize}

\subsection{Neuron importance assessment and omission}
\label{subsec:exp-ns}

\begin{figure*}[t]
	\centering
	\subfigure[Training loss under different neuron importance metrics with the sampling neuron selection method.]{
		\begin{minipage}[b]{0.3\textwidth}
			\includegraphics[width=1\textwidth]{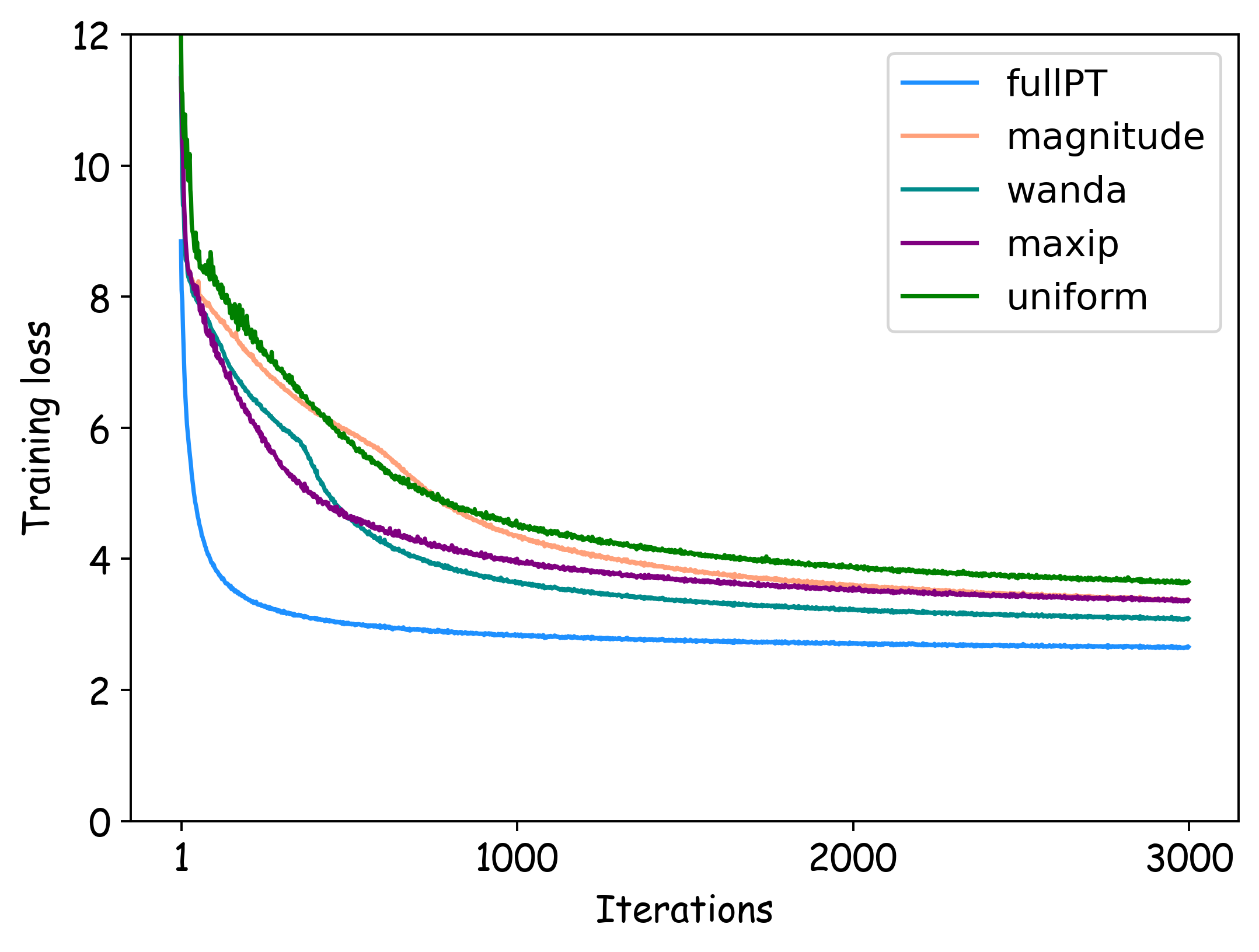}
		\end{minipage}
		\label{fig:nim}
	} \hspace{.05in}
    \subfigure[Traning loss under uniform and maxip with LORS depicted in the right subfigure.]{
    	\begin{minipage}[b]{0.3\textwidth}
   		\includegraphics[width=1\textwidth]{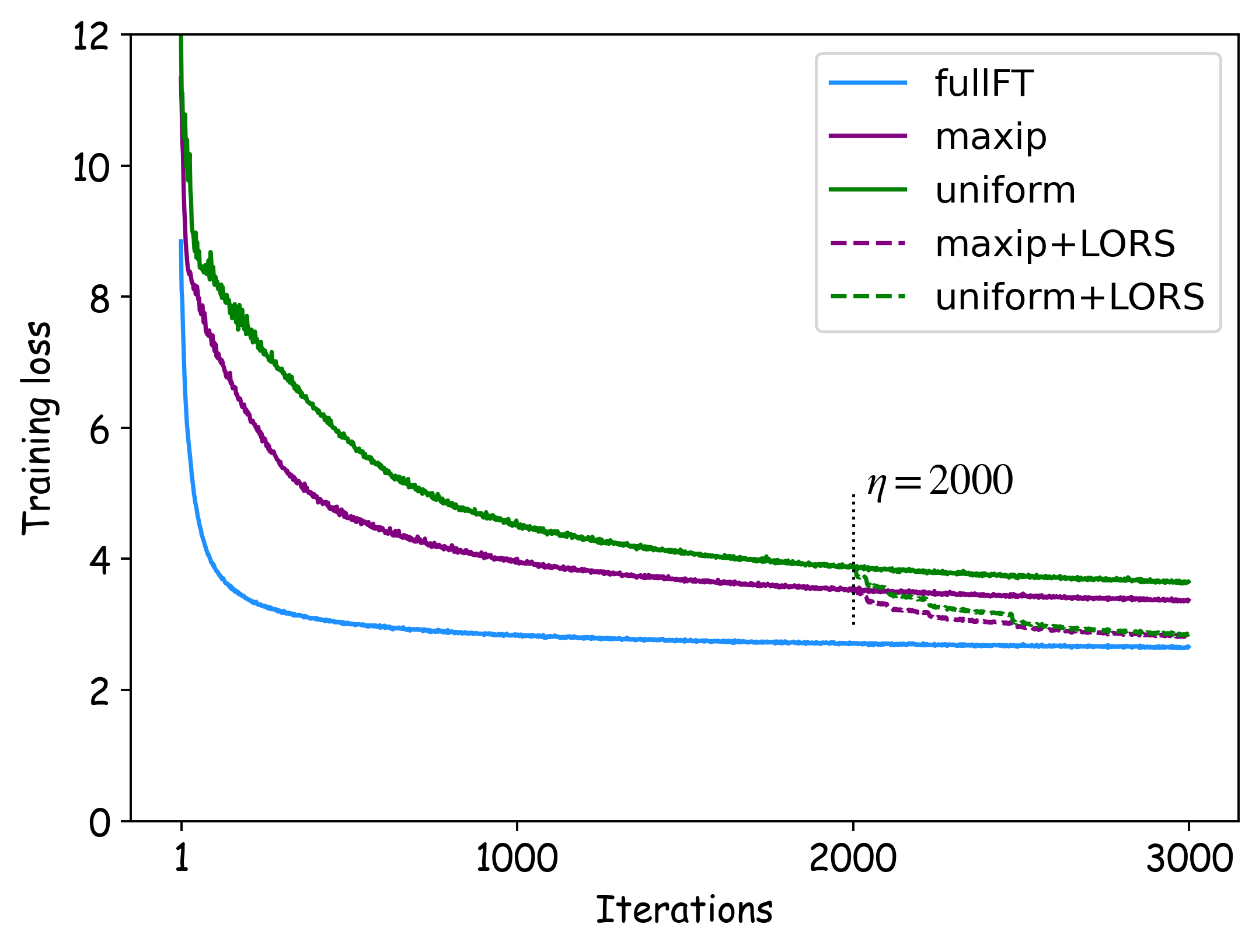}
    	\end{minipage}
	\label{fig:lors}
    } \hspace{.05in}
     \subfigure[The ladder omission rate scheduler~(LORS). $T=3000,\eta=2000,r=0.5.$ ]{
    	\begin{minipage}[b]{0.3\textwidth}
   		\includegraphics[width=1\textwidth]{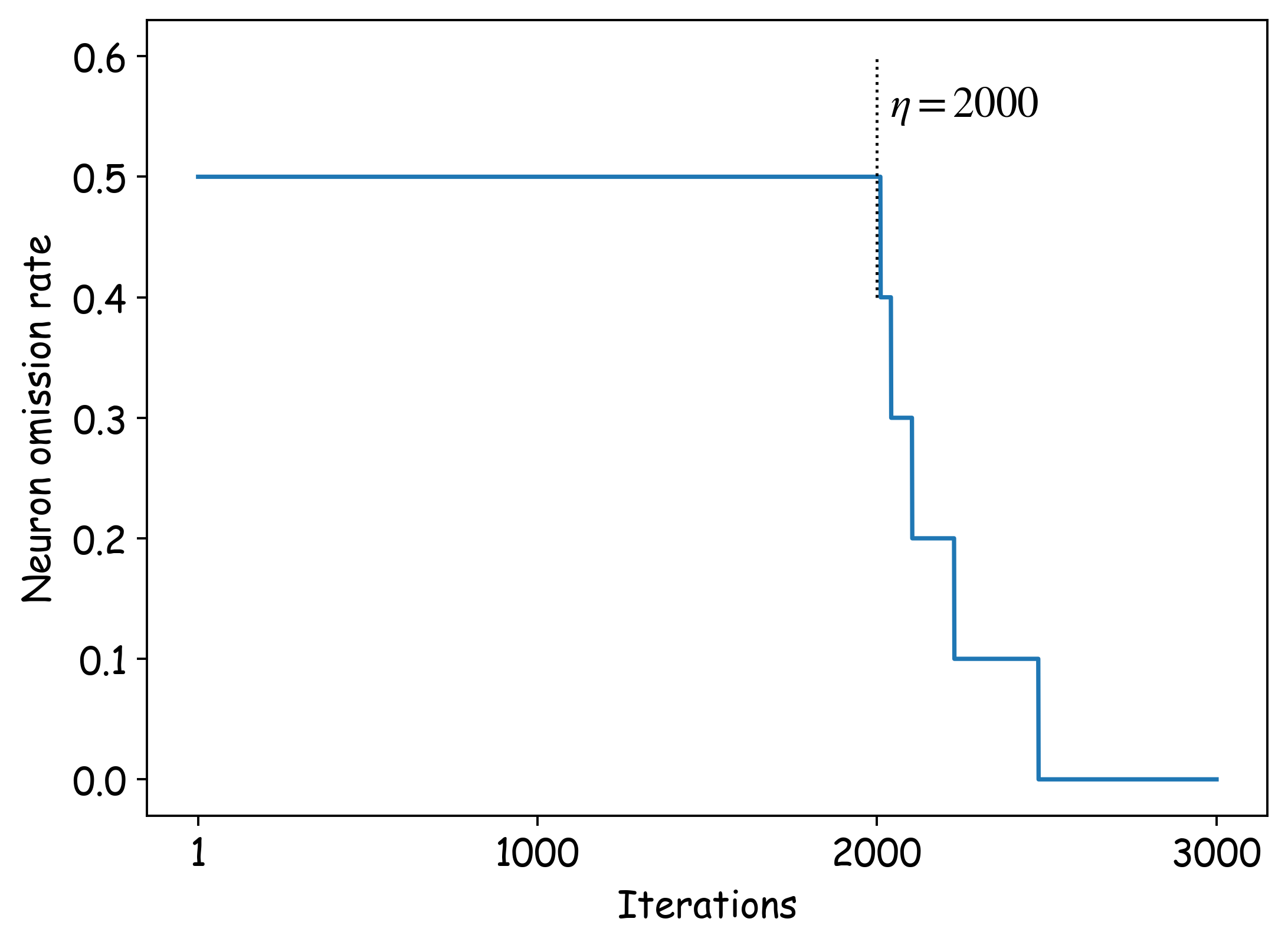}
    	\end{minipage}
	\label{fig:or}
    }
    \setlength{\abovecaptionskip}{0em}
\setlength{\belowcaptionskip}{0em}
	\caption{Continual pre-training of TinyLlama-1.1B}
	\label{fig:CPT-1.1B}

 \vspace{-3mm}
\end{figure*}

We first conduct experiments over relatively small models to answer Q1 due to limitations in computing resources and time.

\paragraph{Setup} We initially continue pre-training TinyLlama-1.1B~\cite{zhang2024tinyllama} over about $12$B tokens of Chinese data to augment its capabilities for understanding and generating Chinese text. Inspired by \citeauthor{cui2023efficient}, we extend its vocabulary with an additional $33000$ Chinese tokens. Following \citeauthor{wei2023skywork}, we validate the improvement of the Chinese language processing capability of models by evaluating perplexity over Skywork~\cite{wei2023skywork}, which is a Chinese language modeling evaluation benchmark across $6$ domains. For neuron importance assessment, we try the four aforementioned metrics: uniform, magnitude~\cite{han2015learning}, wanda~\cite{,sun2023simple} and maxip~\cite{liu2023deja}. For neuron selection methods, we explore both top-k and sampling. All results are obtained by the popular LM-Evaluation-Harness~\cite{eval-harness} tool.

\paragraph{Hyperparameters} We continue pre-training TinyLlama-1.1B by Megatron-DeepSpeed~\cite{smith2022using} framework on $32$ A800-80G GPUs with Zero-3~\cite{rajbhandari2020zero} and FlashAttention~\cite{dao2022flashattention,dao2023flashattention} techniques. The batch size is $1024$ and the maximum length is $4096$. We adopt a cosine learning rate scheduler with a maximum learning rate of $5$e-$5$. For the ladder omission rate scheduler, $T=3000, L=5$, and $\eta=2000$. For the sampling neuron selection method, we explore three different temperature settings: $\left[0.1, 0.05, 0.01\right]$. The results are shown in Table \ref{tab:temperature} of Appendix \ref{sec:appendix1}. Finally, we set $\tau=0.05$ in all our experiments.

\begin{table*}[t]

\centering
\resizebox{\textwidth}{!}{
\begin{tabular}{c|c|cccccc:c|cc:c} 
\hline

\hline
%--------------------------------------------------------
\multirow{2}{*}{\textbf{Order}} & \multirow{2}{*}{\textbf{CPT}} & \multicolumn{6}{c:}{\textbf{Skywork PPL}~$\mathbf{\downarrow}$} & \multirow{2}{*}{\textbf{Avg.}~$\mathbf{\downarrow}$} & \multirow{2}{*}{\textbf{CMMLU}~$\mathbf{\uparrow}$} & \multirow{2}{*}{\textbf{AGI-Eval}~$\mathbf{\uparrow}$} & \multirow{2}{*}{\textbf{Avg.}~$\mathbf{\uparrow}$}  \\ 
\cdashline{3-8}
 & & finance & game & general & government & movie & technology & & & & \\ 
\hline
%--------------------------------------------------------
\rowcolor{green!7} \multicolumn{12}{c}{\emph{Llama-2 7B}}\\
1 & Original & $6.48$ & $12.65$ & $7.79$ & $8.60$ & $18.70$ & $8.81$ & $10.51$ & $32.27$ & $30.04$ & $31.16$ \\
2 & FullCPT & $3.37$ & $8.24$ & $4.50$& $3.77$ & $11.04$ & $5.76$ & $6.11$ & $32.88$ & $31.94$ & $32.41$ \\
3 & SPDF & $3.91$ & $9.35$ & $5.10$ & $4.49$ & $12.79$ & $6.40$ & $7.01$ & $31.89$ & $31.79$ & $31.84$\\
4 & SAT~(ours) & $3.69$ & $8.78$ & $4.80$ & $4.14$ & $11.86$ & $6.07$ & $6.56$  & $32.79$ & $31.84$ & $32.32$\\

\rowcolor{green!7} \multicolumn{12}{c}{\emph{Llama-2 13B}}\\
5 & Original & $6.00$ & $11.46$ & $7.26$ & $7.59$ & $17.22$ & $8.09$ & $9.60$ & $38.08$ & $37.73$ & $37.91$ \\
6 & FullCPT & $3.25$ & $7.88$ & $4.38$ & $3.61$ & $10.70$ & $5.54$ & $5.89$ & $44.53$ & $40.02$ & $42.28$ \\
7 & SPDF & $3.73$ & $8.82$ & $4.89$ & $4.24$ & $12.31$ & $6.08$ & $6.68$ & $40.71$ & $37.20$ & $38.96$ \\
8 & SAT~(ours) & $3.48$ & $8.23$ & $4.58$ & $3.89$ & $11.22$ & $5.76$ & $6.19$ & $44.36$ & $39.78$ & $42.07$\\

\hline

\hline
\end{tabular}
}
\caption{Language modeling and Chinese benchmark evaluation results of SAT with optimal configuration in the continual pre-training scenario. CPT: continual pre-training. SAT: sparsity-accelerated training.}
\label{tab:large}
\end{table*}

\paragraph{Results} First, from Table \ref{tab:ns}~(line 4-6 vs. 7-9), we discover the neuron selection method of sampling beats top-k under all neuron importance metrics~(Q1). We suspect the top-k method tends to concentrate on selecting relatively fixed neurons which causes overfitting. Then, Figure \ref{fig:CPT-1.1B} demonstrates the continual pre-training loss of TinyLlama-1.1B. From Figure \ref{fig:nim}, we can find the lowest losses achieved by all neuron importance metrics are higher than those of full pre-training due to the lack of half-trainable parameters in each iteration. Among all neuron importance metrics, wanda obtains the lowest training loss, while not as smooth as uniform and maxip during early stages~(about 300-500 iterations). Considering the language modeling evaluation results of Table \ref{tab:ns}~(line 3, 7-9), the uniform and maxip equipped with sampling neuron selection method may be reasonable SAT configuration candidates.

Subsequently, we introduce the ladder omission rate scheduler~(see Figure \ref{fig:or}). We can observe the training loss decreases rapidly to a level comparable to full pre-training for both configuration candidates. However, the maxip metric obtains a faster rate of loss reduction. Combining with the results in Table \ref{tab:ns}~(line 10-11), we conclude \emph{maxip} neuron importance metric with \emph{sampling} neuron selection method, plus the \emph{LORS} are optimal configurations for SAT~(Q1). Furthermore, SAT achieves comparable perplexity to full pre-training with such optimal configuration.

\subsection{Performance effects of SAT in both scenarios}
\label{subsec:exp-performance}
Next, we validate whether SAT with the optimal configuration works in both continual pre-training and supervised fine-tuning scenarios~(Q2).

\paragraph{Setup} To better simulate real-world application scenarios, we perform experiments on larger LLMs~($7$B and $13$B). For the continual pre-training scenario, we adopt the identical pre-training setup to TinyLlama-1.1B. Besides the language modeling evaluation on Skywork, we evaluate LLMs on two additional popular Chinese benchmarks: CMMLU~\cite{li2023cmmlu} and AGI-Eval~\cite{zhong2023agieval}\footnote{We only evaluate on single-choice problems.}. For the supervised fine-tuning scenario, we instruction-tune LLMs on $50$K subsample of Flan v2~\cite{10.5555/3618408.3619349,wang2023far}. Following \citeauthor{ivison2023camels}, we assess the instruction-tuned LLMs on $\textsc{GPT4All}$~\cite{anand2023gpt4all} for reasoning ability and MMLU~\cite{hendrycks2020measuring} for factuality ability.

\begin{table*}[t]
\centering
\resizebox{\textwidth}{!}{
\begin{tabular}{c|cl|ccccccc:c|c} 
\hline

\hline
%--------------------------------------------------------
\multirow{2}{*}{\textbf{Order}} & \multirow{2}{*}{\textbf{SFT}} & \multirow{2}{*}{\textbf{Method}} & \multicolumn{7}{c:}{\textbf{\textsc{GPT4All} Acc}~$\mathbf{\uparrow}$} & \multirow{2}{*}{\textbf{Avg.}~$\mathbf{\uparrow}$} & \multirow{2}{*}{\textbf{MMLU}~$\mathbf{\uparrow}$} \\ 
\cdashline{4-10}
 & & & HellaSwag & Obqa & WinoGrande & $\text{ARC}_\text{c}$ & $\text{ARC}_\text{e}$ & boolq & piqa & \\ 
\hline
%--------------------------------------------------------
\rowcolor{gray!10} \multicolumn{12}{c}{\emph{Llama-2 7B}}\\
1 & \XSolidBrush & Original & $58.30$ & $36.40$ & $73.80$ & $49.23$ & $79.29$ & $79.36$ & $78.78$ & $65.02$ & $45.8$  \\
2 & \Checkmark & FullFT & $\mathbf{62.79}$ & $40.00$ & $74.82$ & $49.32$ & $79.17$ & $84.01$ & $77.80$ & $66.84$ & $50.5$ \\
\cdashline{1-12}
3 & \Checkmark & $\left(\text{IA}\right)^3$ & $58.46$ & $36.40$ & $74.35$ & $48.46$ & $79.63$ & $80.52$ & $79.27$ & $65.30$ & $46.7$\\
4 & \Checkmark & SpIEL & $58.74$&	$37.40$&	$75.30$	&$52.30$&	$81.44$	&$\mathbf{84.37}$	&$79.38$&	$66.99$	&$\mathbf{50.7}$ \\
5 & \Checkmark & LoRA & $58.06$  & $37.80$ & $75.30$ & $50.09$ & $79.88$ & $83.27$ & $79.43$ & $66.26$ & $49.3$ \\
6 & \Checkmark & SAT~(ours) & $59.71$ & $\mathbf{40.20}$ & $\mathbf{76.01}$ & $\mathbf{52.56}$ & $\mathbf{81.52}$ & $83.85$ & $\mathbf{79.71}$ & $\mathbf{67.65}$ & $50.5$ \\

\rowcolor{gray!10} \multicolumn{12}{c}{\emph{Llama-2 13B}}\\
7 & \XSolidBrush & Original & $61.27$ & $37.20$ & $77.11$ & $52.82$ & $82.03$ & $83.70$ & $79.38$ & $67.64$ & $55.3$ \\
8 & \Checkmark & FullFT & $\mathbf{65.86}$ & $40.40$ & $74.66$ & $52.22$ & $80.93$ & $83.73$ & $79.00$ & $68.11$ & $55.6$ \\
\cdashline{1-12}
9 & \Checkmark & SpIEL & $61.15$&	$38.40$&	$78.45$&	$54.27$&	$82.95$&	$86.45$&	$79.98$&	$68.81$	&$\mathbf{55.8}$\\
10 & \Checkmark & $\left(\text{IA}\right)^3$ & $61.62$ & $36.60$ & $77.11$ & $53.16$ & $82.28$ & $85.11$ & $79.98$ & $67.98$ & $54.3$ \\
11 & \Checkmark & LoRA & $61.01$ & $40.00$ & $78.61$ & $52.05$ & $82.07$ & $86.06$ & $80.20$ & $68.57$ & $\mathbf{55.8}$ \\
12 & \Checkmark & SAT & $62.29$ & $\mathbf{41.60}$ & $\mathbf{78.85}$ & $\mathbf{56.57}$ & $\mathbf{83.12}$ & $\mathbf{86.82}$ & $\mathbf{81.18}$ & $\mathbf{70.06}$ & $55.4$ \\

\hline

\hline
\end{tabular}
}

\caption{Factuality and Reasoning evaluation results of SAT with optimal configuration in the supervised fine-tuning scenario. SFT: supervised fine-tuning. $\left(\text{IA}\right)^3$~\cite{liu2022few}, SpIEL~\cite{ansell2024scaling} and LoRA~\cite{hu2022lora} are popular PEFT methods for SFT. FullFT: full fine-tuning.}
\label{tab:sft}
\end{table*}

\paragraph{Hyperparameters} For the continual pre-training, all pre-training hyperparameters are the same as those of TinyLlama-1.1B. We evaluate models on CMMLU and AGI-Eval in a $3$-shot manner. For the supervised fine-tuning, we adopt the Huggingface+DeepSpeed~\cite{wolf2019huggingface,rasley2020deepspeed} framework with Zero-3 technique. We instruction-tune models $3$ epochs with a cosine learning rate scheduler whose maximum learning rate is $1$e-$5$ on $64$ A800-80G GPUs. The batch size is $128$. Similar to PEFT methods~\cite{hu2022lora,liu2022few}, we only tune a small proportion of parameters at each training step. In particular, we set $r=96\%$, $T=1200, \eta=600$, and $L=1$. The models are evaluated in a $5$-shot manner for both \textsc{GPT4All} and MMLU. More details of baselines are in Appendix \ref{sec:appendix2}.

\begin{table*}[t]
\centering

\resizebox{\linewidth}{!}{
\begin{tabular}{cc|cc:ccc:ccc} 
\hline

\hline

% ------------------------------------------------------------------
\multirow{2}{*}{\textbf{Scenario}}  & \multirow{2}{*}{\textbf{Model}} & \multicolumn{2}{c:}{\textbf{Saving}~(\%)} & \multicolumn{3}{c:}{\textbf{7B}} & \multicolumn{3}{c}{\textbf{13B}} \\
\cdashline{3-10}
& & FLOPs & FLOPS~($\downarrow$) & Time~(h) & Speedup~(\%) & PM~(GB) & Time~(h) & Speedup~(\%) & PM~(GB) \\

\hline

\multirow{2}{*}{CPT} & FullCPT & - & - & $36.5$ & - & $44.7$ & $66.4$ & - & $58.4$ \\
& SAT~(ours) & $33$ & $6$ & $26.1$ & $28.5$ & $44.7$ & $45.9$ & $30.9$ & $58.3$\\

\hline
\multirow{4}{*}{SFT} & FullFT & - & - & $1.8$ & - & $23.9$ & $3.1$ & - & $36.6$ \\
& SpIEL & - & - & $1.5$ & $16.7$ & $10.3$ & $3.0$ & $3.2$ & $13.5$\\
& LoRA & - & - & $1.2$ & $33.3$ & $8.1$ & $2.7$ & $12.9$ & $13.1$\\
& SAT~(ours) & $48$ & $16$ & $1.1$ & $38.9$ & $23.1$ & $2.0$ & $35.5$ & $36.4$ \\

\hline

\hline
\end{tabular}
}
\caption{Efficiency of SAT on Llama-2 7B and 13B for both continual pre-training~(CPT) and supervised fine-tuning~(SFT). FLOPs:floating operations; FLOPS: floating operations per second; PM: Peak Memory.}
\label{tab:time}
\end{table*}

\paragraph{Results} Table \ref{tab:large} shows the results of continual pre-training. First, the continual pre-training enhances the Chinese text-processing ability of LLMs~(line $1$ vs. $2$; line $5$ vs. $6$). The models perform comparably with SAT to full pre-training~(line $2$ vs. $4$; line $6$ vs. $8$). Furthermore, we adjusted to adapt SPDF~\cite{thangarasa2023spdf} for continual pre-training. Specifically, we conducted sparse pre-training first, followed by dense continual pre-training with the model. We can find our SAT consistently outperforms SPDF~(line $3$ vs. $4$; line $7$ vs. $8$). We can state the SAT harms little to the performance of LLMs in the continual pre-training scenario~(Q2).

Table \ref{tab:sft} illustrates the results of supervised fine-tuning. The SFT boosts the performance of Llama-2 7B a lot~(line $1$ vs. $2$) while contributing minor improvements to Llama-2 13B~(line $7$ vs. $8$). We attribute this to the fact that Llama-2 13B, with its greater number of parameters, has a relatively strong inherent capacity before SFT~(line $1$, $2$, and $7$). Maybe it requires more SFT data to achieve further performance improvements. Compared to famous PEFT methods, $\left(\text{IA}\right)^3$, SpIEL, and LoRA, our SAT obtains better performance on \textsc{GPT4All} and comparable results on MMLU. Furthermore, our SAT achieves comparable results on MMLU and even better accuracy on \textsc{GPT4All} compared to FullFT. We suspect that this may be due to the removal of some unnecessary neurons, which alleviates the risk of overfitting. Overall, the SAT has almost no bad effects on the model performance~(Q2).

After confirming that SAT has little to no detrimental effect on model performance, we finally examine the actual acceleration it can achieve~(Q3).

\begin{figure}[t]
	\centering
	\subfigure[Throughput of continual pre-training.]{
		\begin{minipage}[b]{0.42\linewidth}
			\includegraphics[width=1\textwidth]{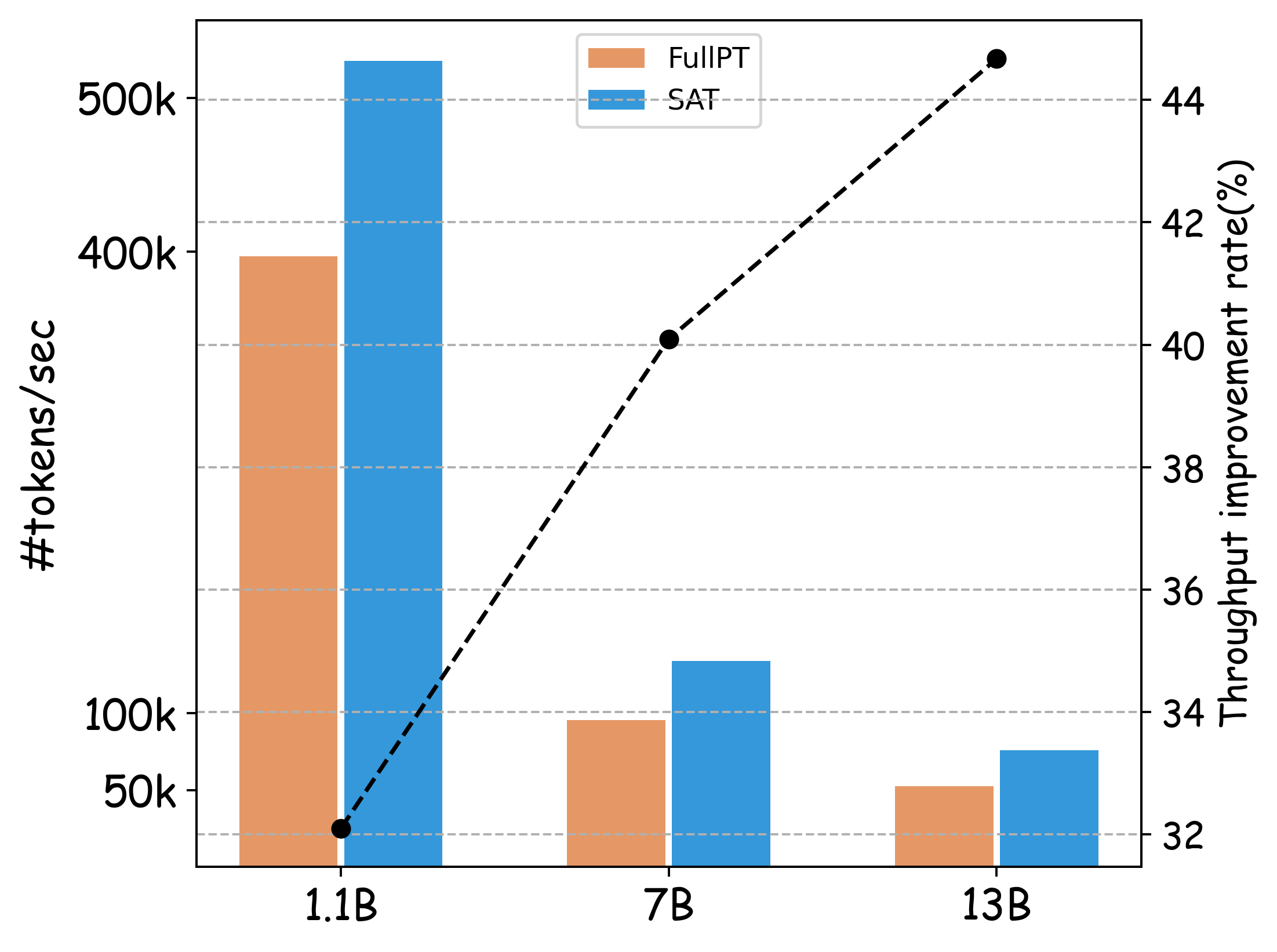}
		\end{minipage}
		\label{fig:cpt}
	} 
    \subfigure[Elapsed time of supervised fine-tuning.]{
    	\begin{minipage}[b]{0.42\linewidth}
   		\includegraphics[width=1\textwidth]{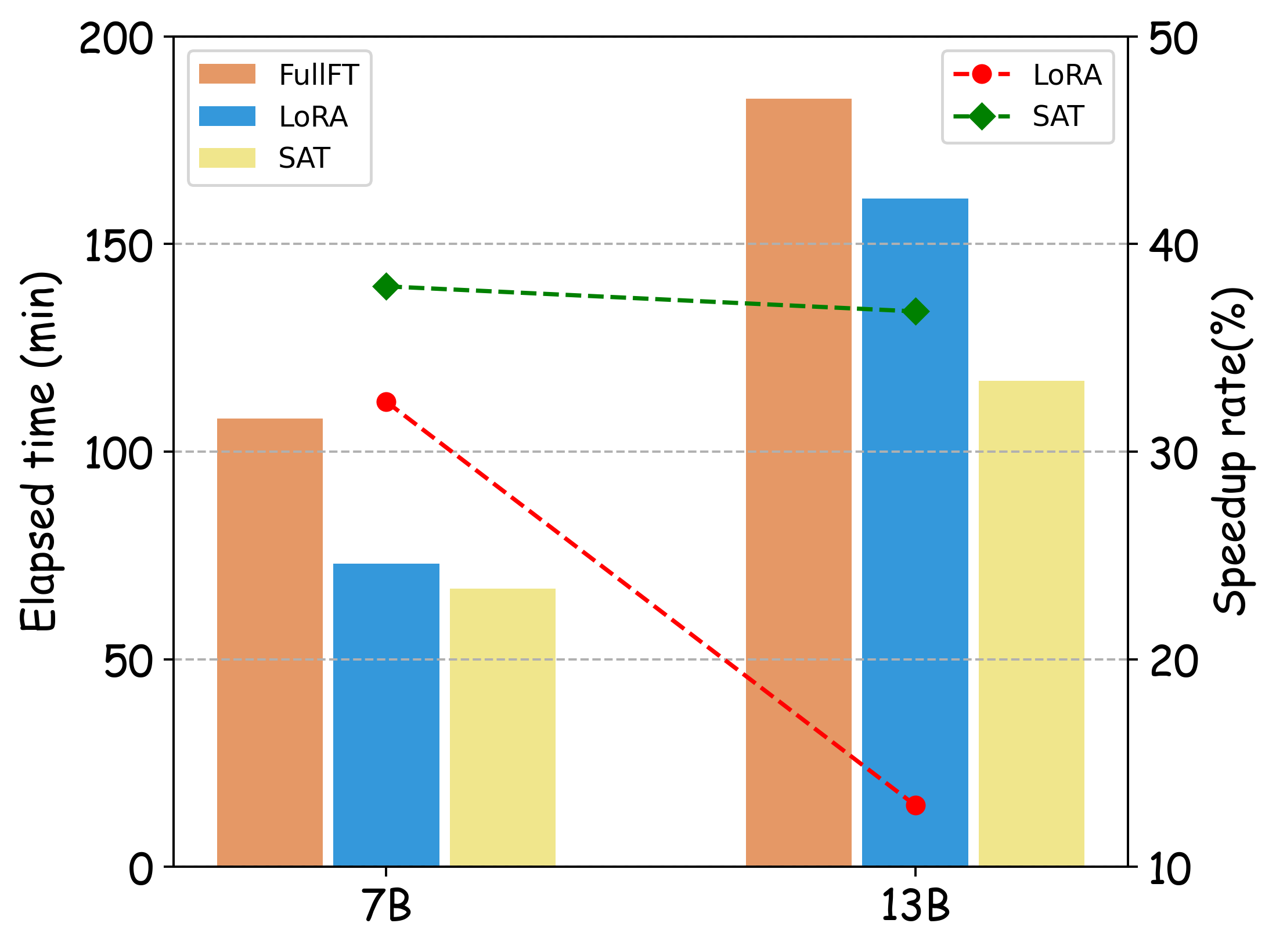}
    	\end{minipage}
	\label{fig:sft}
    } 
    \caption{Speedup of SAT for both continual pre-training and supervised fine-tuning.}
    \label{fig:speedup}
    
\end{figure}

\subsection{Efficiency of SAT}
\label{subsec:exp-speedup}

\paragraph{Setup} We compare the speedup of SAT with standard training in both continual pre-training and supervised fine-tuning scenarios. All experiments are performed on identical $4$ machines, each equipped with $8$ A800 GPUs under the setup described in \cref{subsec:exp-ns} and \cref{subsec:exp-performance}. For the continual pre-training, we compare the number of processed tokens per second since all training samples are truncated or padded to the same length. For the supervised fine-tuning, we compare the total elapsed training time due to the variable lengths of training samples.

\paragraph{Results} Figure \ref{fig:speedup} provides the speedup results for both continual pre-training and supervised fine-tuning. For the continual pre-training, we discover SAT improves throughput increases with the growth of model sizes~(the dashed line in Figure \ref{fig:cpt}) and the maximum throughput improvement approaches $45\%$. For the supervised fine-tuning, the speedup of LoRA decreases significantly when the model size is relatively large. However, our SAT is more stable. In general, SAT can obtain about $38\%$ speedup in elapsed time. 

Table \ref{tab:time} presents the results in terms of theoretical FLOPs, actual FLOPS, training time, and memory usage. The FLOPS decrease is attributed to the reduced computation required for each iteration. As we can see, SAT achieved the optimal acceleration effect; however, there was no improvement in memory usage, which is consistent with the theoretical analysis and experimental results in the previous discussion. Finally, we can determine SAT indeed accelerates both continual pre-training and supervised fine-tuning~(Q3).
\label{sec:related-work}
\section{Related Works}
\paragraph{Transformer Pruning} Pruning has demonstrated considerable potential in augmenting the inference speed and reducing the memory footprint of transformers, as highlighted in prior research~\cite{kwon2022fast}. Broadly speaking, transformer pruning techniques can be delineated into two principal categories~\cite{chitty2023survey}: unstructured pruning and structured pruning. Unstructured pruning methods~\cite{gordon2020compressing,campos2022sparse,zhang2022platon} possess the capability to substantially diminish the number of parameters due to their fine-grained pruning granularity. Nonetheless, they frequently encounter challenges in achieving significant improvements in inference speed across diverse hardware platforms~\cite{sanh2020movement}.

Hence, numerous researchers are embracing structured pruning methods at the granularity level of entire layers, filters, channels, or heads~\cite{li2021differentiable}. For instance, \citeauthor{liu2021ebert} prunes the unimportant heads within multi-head attention layers and insignificant channels in feed-forward networks on BERT~\cite{kenton2019bert}. \citeauthor{liu2023deja} and \citeauthor{song2023powerinfer} extend such a method and scale it for large language models. In this study, distinct from their focus on the inference of large language models~(LLMs), we draw upon the principles of structured pruning and apply them to the training of LLMs.

\paragraph{Sparse Fine-Tuning}
Sparse fine-tuning is a technique aimed at expediting model fine-tuning and diminishing model memory demands by selectively updating only a small subset of parameters~\cite{ansell-etal-2022-composable,anonymous2024evolving}. For relatively small language models, such as BERT, \citeauthor{guo-etal-2021-parameter} devise a binary mask controlled by a regularization term to manage the parameters for updating. \citeauthor{sung2021training} induce the mask by retaining parameters exhibiting higher Fisher information over numerous iterations and subsequently maintain the mask to sparsely fine-tune models. Additionally, \citeauthor{ansell-etal-2022-composable} retain parameters that undergo the most significant changes during an initial round of full fine-tuning.

With the widespread adoption of LLMs, \citeauthor{ansell2024scaling} and \citeauthor{zhao2024apt} commence scaling sparse fine-tuning to LLMs. However, to mitigate the memory overhead of LLMs, they both incorporate the parameter-efficient fine-tuning (PEFT) methods~\cite{houlsby2019parameter,li-liang-2021-prefix,lialin2023scaling}, which may not always be entirely suitable~\cite{cui2023efficient,zhao2024chemdfm}. In this work, instead of employing the PEFT methods, we place greater emphasis on conventional sparse training techniques, making it applicable to common scenarios such as continual pre-training and supervised fine-tuning for LLMs.

\section{Conclusion}
In this paper, we aim to accelerate two regular additional training scenarios of LLMs, i.e., continual pre-training and supervised fine-tuning. Our main intuition is to leverage the sparsity of neurons in large models, as discovered by previous researchers, to accelerate training by disregarding computations for unimportant neurons and propose a sparsity-accelerated training~(SAT) framework. Extensive experiments on several benchmarks demonstrate SAT can accelerate the training of LLMs and maintain performance simultaneously.
\section*{Limitations}
On one hand, this work primarily focuses on large language models and can explore even larger and more diverse model structures in the future. Furthermore, several other neuron importance metrics are necessary to explore. On the other hand, training numerous large-scale models requires substantial computational resources, incurring high costs and resulting in significant carbon emissions.

\section*{Acknowledgements}
We thank all the reviewers for their valuable suggestions on this work. This work is funded by the China NSFC Projects (92370206, U23B2057,62106142 and 62120106006) and Shanghai Municipal Science and Technology Major Project (2021SHZDZX0102).

\bibliography{reference}

\appendix
\section{Temperature Exploration }
\label{sec:appendix1}
\begin{table}[htbp]
\centering

\resizebox{\linewidth}{!}{
\begin{tabular}{c|cccccc:c} 
\hline

\hline

% ------------------------------------------------------------------
\multirow{2}{*}{Temperature}  & \multicolumn{6}{c:}{\textbf{Skywork PPL}~$\mathbf{\downarrow}$} & \multirow{2}{*}{\textbf{Avg.}~$\mathbf{\downarrow}$}     \\ 
\cdashline{2-7}
 &  \multicolumn{1}{c}{finance} & \multicolumn{1}{c}{game} & \multicolumn{1}{c}{general} & \multicolumn{1}{c}{government} & \multicolumn{1}{c}{movie} & \multicolumn{1}{c:}{technology} &  \\ 
\hline
% ------------------------------------------------------------------
$0.1$ & $6.71$ & $18.05$ & $8.74$ & $8.32$ & $25.68$ & $11.24$ & $13.12$ \\
$0.05$ & $\mathbf{6.56}$ & $\mathbf{17.58}$ & $\mathbf{8.59}$ & $\mathbf{8.10}$ & $\mathbf{24.89}$ & $\mathbf{11.03}$ & $\mathbf{12.79}$ \\
$0.01$ & $9.87$ & $23.16$ & $11.51$ & $14.06$ & $34.90$ & $14.74$ & $18.04$ \\
   
% ------------------------------------------------------------------
\hline

\hline
\end{tabular}
}
\caption{Perplexity scores on the Skywork benchmark of TinyLLaMA (1.1B) after continual pre-training with different temperature settings with the maxip neuron importance metric}
\label{tab:temperature}
\end{table}

\section{Baselines Hyperparameters in Supervised Fine-tuning}
\label{sec:appendix2}
\begin{table}[htbp]
\centering

\resizebox{\linewidth}{!}{
\begin{tabular}{c|ccc|ccc} 
\hline

\hline

% ------------------------------------------------------------------
\multirow{2}{*}{Model}  & \multicolumn{3}{c|}{7B} & \multicolumn{3}{c}{13B}     \\ 

\cdashline{2-7}

& lr & $r$ & $\lambda$ & lr & $r$ & $\lambda$ \\

\hline

\hline

$\text{(IA)}^3$ & 	3e-4 & - & - & 1e-4 & - & -\\
LoRA & 1e-4	& 64 & - & 	3e-5&	64&	- \\
SpIEL	&1e-4&	64	&30 & 1e-5&	64&	30 \\
FullFT	&1e-5	&-	&-&1e-5	&-	&-\\
SAT~(ours)	&1e-5	&-	&-&1e-5	&-	&-\\

\hline

\hline
\end{tabular}
}
\caption{Baselines hyperparameters in supervised fine-tuning for LLaMA-2 7B and 13B. lr: learning rate; $r$: rank in LoRA; $\lambda$: weight decay strength in SpIEL}
\label{tab:baseline_hyp}
\end{table}

\end{document}